\pdfoutput=1

\documentclass[11pt]{article}

\usepackage{times}
\usepackage{latexsym}

\usepackage[T1]{fontenc}

\usepackage[utf8]{inputenc}

\usepackage{microtype}

\usepackage{inconsolata}

\usepackage{graphicx}

\usepackage{amsmath}
\usepackage{amssymb}
\usepackage{mathtools}
\usepackage{amsthm}

\usepackage{hyperref}

\usepackage[shortlabels]{enumitem}

\newcommand{\RR}{\mathbb{R}}

\renewcommand{\t}{\mathbf{t}}

\newcommand{\D}{\mathcal{D}}

\newcommand{\T}{\mathcal{T}}

\title{`Generalization is hallucination'\\through the lens of tensor completions}

\author{Liang Ze Wong \\
  \texttt{liangze.wong@gmail.com} \\}

\begin{document}
\maketitle
\begin{abstract}
In this short position paper, we introduce tensor completions and artifacts and make the case that they are a useful theoretical framework for understanding certain types of hallucinations and generalizations in language models.

\end{abstract}

\section{Introduction}
Generalization and hallucination in generative language models are often studied independently, the former as a feature to be encouraged, the latter as a bug to be avoided.
In this short position paper, which expands upon Section 2.1 of \cite{wong2025paying}, we introduce tensor completions, and make the case that both generalizations and hallucinations arise as tensor completion artifacts.

We begin by defining tensor completions in Section \ref{sec:tensor_completions_and_artifacts}, relating them to language models, and showing how they give rise to completion artifacts, which are novel sentences that a model predicts with high probability.
With experiments on toy models and datasets, we show that artifacts are prevalent, and increase in number when models are smaller.

The value of a theoretical framework lies in its ability to explain and organize phenomena and suggest lines of research, and we demonstrate this in Section \ref{sec:generalization_or_hallucination}, where we discuss the implications of our framework on generalizations, hallucinations and overfitting, and relate our findings to the existing literature.
We also call for future work on mitigating hallucinations to also consider the effects on generalization, and vice versa.
We end by listing the limitations of this paper and directions for future work in Section \ref{sec:limitations_and_future_directions}.

\section{Tensor completions and artifacts}
\label{sec:tensor_completions_and_artifacts}

\begin{figure*}[t]
  \includegraphics[width=0.25\linewidth]{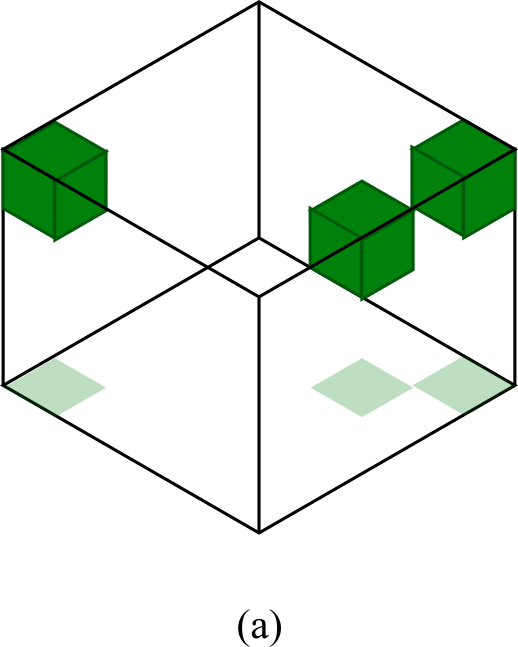} \hfill
  \includegraphics[width=0.25\linewidth]{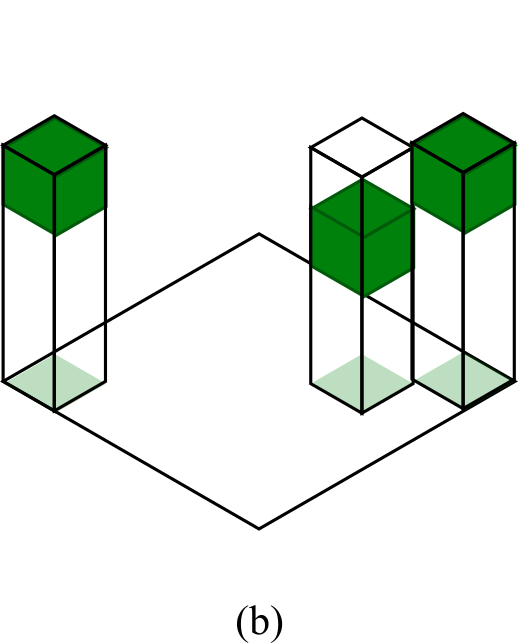}\hfill
  \includegraphics[width=0.25\linewidth]{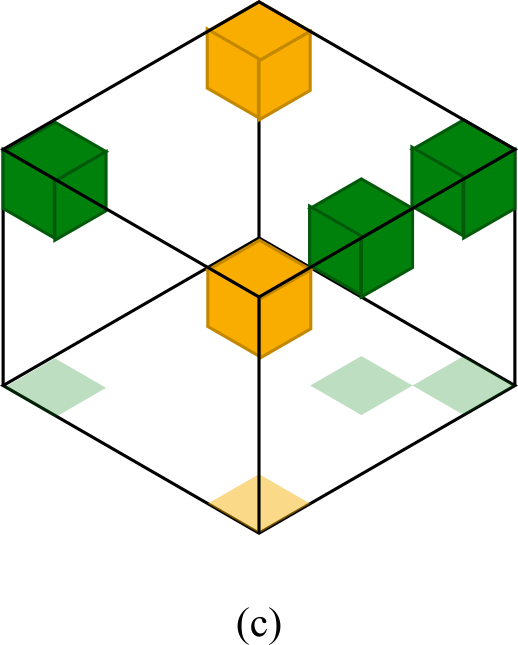}
  \caption{Illustration of how tensor completion can give rise to new sentences, or \emph{artifacts}. (a) A tensor $D^3$ associated to a corpus. Each green box is $1$, and represents a sentence in the corpus. Empty space is $0$.  The horizontal plane is the $(t_1, t_2)$-plane, while the vertical axis is $t_3$. (b) Fibers of $D^3$ that are `seen' by a language model during training. (c) A low-rank completion $D'$ that is consistent with the training fibers, but  also has additional artifacts (i.e. new sentences, colored orange). The rank of the completion $D'$ is 2, while the original $D^3$ has rank 3.}
  \label{fig:tensors}
\end{figure*}

\begin{figure}[t]
  \includegraphics[width=\columnwidth]{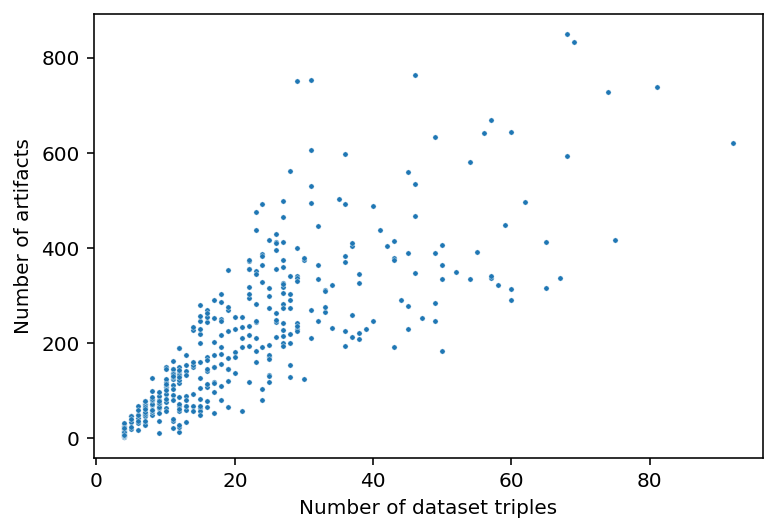}
  \caption{Number of artifacts against number of triples in the training dataset, over 400 random datasets. Artifacts are triples $(t_1, t_2, t_3)$ which are not in the dataset, but the model still predicts $t_3$ with high probability ($\geq 0.95$) when given $(t_1,  t_2)$. The same attention-only model with $n_{layers} = 1, n_{head} = 4, d_{model} = 8$ and $d_{head} = 2$ was used throughout.}
  \label{fig:new_triples_vs_num_triples}
\end{figure}

\begin{figure}[t]
  \includegraphics[width=\columnwidth]{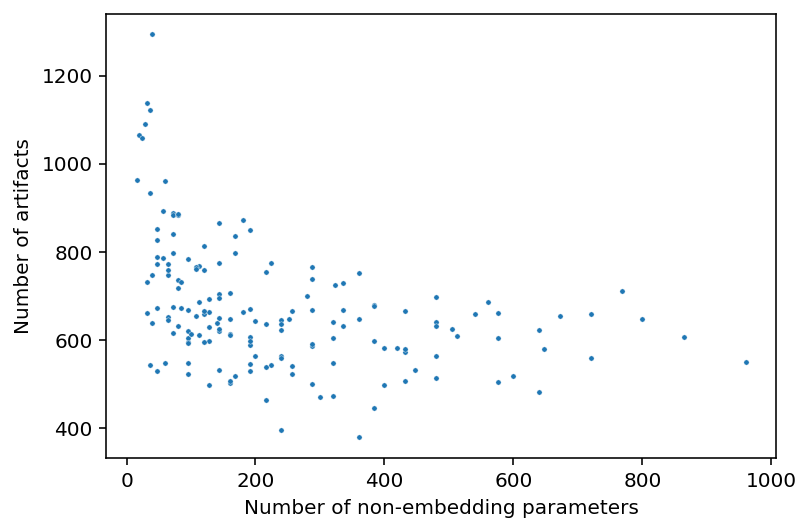}
  \caption{Number of artifacts against number of non-embedding parameters in 165 models with $n_{head} \leq 4, d_{model} \leq 10$ and $d_{head} \leq 6$. The same dataset with 29 triples and 44 tokens was used, so the maximum possible number of artifacts is $44^2 - 29 = 1,907$ .}
  \label{fig:new_triples_vs_num_params}
\end{figure}

We assume we have a fixed set of tokens $\T$, a maximum context size $N$, and a corpus or dataset $\D$ consisting of sentences or tuples $\t = (t_1, t_2, \dots, t_n)$ where $t_i \in \T$ and $n \leq N$.
A common loss function for generative language models such as transformers \cite{vaswani2017attention} is 
\begin{equation}
\label{eq:loss}
  \min_{\theta} \sum_{\t \in \D}\sum_{i}^{|\t|} -\log(P_\theta(t_i | t_1, \dots, t_{i-1} ))
\end{equation}
where $P_\theta$ is the next token distribution from a model with weights $\theta$.
For our purposes, it will be convenient to define collections of $n$-truncated sentences $\D^n := \{(t_1, \dots, t_n) \,|\, (t_1, \dots, t_n, \dots, t_\ell) \in \D\} \subset \T^n$ and  rewrite (\ref{eq:loss}) as 
\begin{equation}
\label{eq:loss_n}
  \min_{\theta} \sum_{n = 1}^N \sum_{\t \in \D^n} -\log(P_\theta(t_n | t_1, \dots, t_{n-1} )).
\end{equation}
This allows us to treat each $n \in N$ separately. 
We define an $n$-dimensional tensor $D^n \in (\RR^{|\T|})^{\otimes n}$ whose 
entries $D^n_{t_1, \dots, t_n} \geq 0$ are the number of times the sentence $(t_1, \dots, t_n)$ appears in $\D^n$ 
(see Fig.~\ref{fig:tensors}a for an example).

The \emph{fiber} of $D^n$ at $\t \in \D^{n-1}$ is a vector (Fig.~\ref{fig:tensors}b) that will be denoted $D^n_{\t, :}$, where the colon `:' indicates that the last index runs over all $t \in \T$.
We say that an $n$-tensor $D'$ is \emph{consistent} with $\D^n$ if $D'_{\t, :} = D^n_{\t, :}$ for all $\t \in \D^{n-1}$.
In practice, having equality is rare, so we allow ourselves to say that $D'$ is consistent as along as the fibers are close according to some similarity measure.
Of course, the original $D^n$ is consistent with $\D^n$, but there can be many more tensors that are consistent.
We define a \emph{completion} of $\D^n$ to be any tensor $D'$ that is consistent with $\D^n$ (Fig.~\ref{fig:tensors}c).

Evidently, language models give rise to completions\footnote{Tensor completion usually refers to algorithms that explicitly complete a tensor, e.g. \cite{ji2019survey,song2019tensor}, but here we take a broader view, allowing algorithms that \emph{implicitly} complete tensors, such as language models. See \cite{wong2025paying} for an early attempt at relating single-layer attention-only transformers to tensor completion.} when they achieve low training loss:
for each model with parameters $\theta$, we may form an $n$-tensor $P^n_\theta$ with entries $(P^n_\theta)_{\t,t_n} = P_\theta(t_n| \t)$  for all $\t \in \T^{n-1}$;
a low loss in (\ref{eq:loss_n}) then implies that the KL divergence from the model distribution $(P^n_\theta)_{\t,:}$ to the data distribution $D^n_{\t, :}$ are low for all $\t \in \D^{n-1}$ i.e. $P^n_\theta$ is consistent with $\D^n$.

Note that $(P^n_\theta)_{\t,:}$ is defined for \emph{all} $\t \in \T^{n-1}$, but we only check consistency on $\t \in \D^{n-1}$.
Indeed, the fibers at $\t \in D^{n-1}$ are the only parts of $D^n$ that the model `sees' during training (Fig.~\ref{fig:tensors}b).
In practice, it is unfeasible to compute $P^n_\theta$ for large $n$ and $\T$, as doing so requires evaluating the model at all possible $(n-1)$-grams.
But \emph{theoretically}, every language model contains the data to compute $P^n_\theta$.

\subsection{Completion artifacts}
Our key observation is that tensor completions give rise to \emph{artifacts}, which are large, non-zero entries in the tensor completion $D'$ that do not belong in $D^n$.
Recall that each cell of $D'$ represents an $n$-gram, and the value of the cell reflects\footnote{Some normalization might need to be done to ensure they are actual probabilities, but heuristically, the larger the value of the cell is, the higher the probability of seeing that $n$-gram.} the probability of the $n$-gram.
So artifacts are \emph{novel $n$-grams or sentences of which the model is very certain, but that do not belong in the training data}.

We have already alluded to one reason for why tensor completions $P^n_\theta$ give rise to artifacts:
models are only trained on a (very small) subset of the entries of $D^n$, namely the fibers $D^n_{\t,:}$ for $\t \in \D^{n-1}$, and the only zeros that are enforced are those that belong to these training fibers.
All other zeros in $D^n$ are not enforced or `seen' by the model, and so there is no reason why we should expect them to be close to zero.
Indeed, language models can produce very large numbers of artifacts, as Fig.~\ref{fig:new_triples_vs_num_triples} shows: we trained a toy model\footnote{Our toy models are simplified versions of the original transformer architecture \cite{vaswani2017attention} without layer normalization, biases or positional encodings. They are commonly used for pedagogical or expositional reasons, e.g. \cite{elhage2021mathematical,nichani2024understanding,wong2025paying}.} on random datasets of triples and counted the number of artifacts, which we defined as triples $(t_1,t_2,t_3) \notin \D^3$ where $P_\theta(t_3| t_1, t_2) \geq 0.95$.
We see that the number of artifacts can be very large, anywhere from 5 to 20 \emph{times} the number of training triples.

Another related reason for artifacts is that language models tend to have constraints on the rank\footnote{The rank of an $n$-tensor $D$ is the minimum number of simple tensors $\mathbf{v_1} \otimes \mathbf{v_2} \otimes \dots \otimes \mathbf{v_n}$ that need to be summed up to give $D$. See e.g. \cite{kolda2009tensor} for a more detailed desciption of tensor rank and its properties.} of their tensor completions (due to constraints on the dimensions of the model weights, for example), and non-zero entries (i.e. artifacts) tend to emerge when attempting to reduce rank.

Fig.~\ref{fig:tensors}c illustrates this with a completion $D'$ that is consistent with Fig.~\ref{fig:tensors}b, but that has more non-zero entries (i.e. artifacts, colored orange) and lower rank than the original tensor in Fig.~\ref{fig:tensors}a.
We see this also in Fig.~\ref{fig:new_triples_vs_num_params}, where we trained toy models with varying numbers of non-embedding parameters\footnote{Number of non-embedding parameters is a common measure of model capacity, e.g. in \cite{lu2024scaling}.} on the same dataset, and counted the number of artifacts produced.
As we reduce the number of parameters, which indirectly reduces rank, the number of artifacts \emph{increases}.

Note that even with large numbers of parameters (and hence a high rank allowance for the tensor completion), the number of artifacts remains high (more than 10 times the number of triples), suggesting that while low rank encourages artifacts, it is not a necessary condition for their occurence.

\section{Generalization or hallucination?}
\label{sec:generalization_or_hallucination}
So tensor completion gives rise to large numbers of artifacts.
How should we treat these artifacts?
Are they generalizations or hallucinations?

Suppose a language model is trained on the following sentences:
\begin{equation}
\label{eq:cheese}
\begin{tabular}{r l l}
  mouse & eats & cheese \\
  mice & eat & cheese
\end{tabular}  
\end{equation} 
The associated tensor $D^3$ would look like the top layer of the tensor in Fig.~\ref{fig:tensors}a, but the top layer of the tensor in Fig.~\ref{fig:tensors}c is also consistent with the data.
The orange boxes would correspond to a language model completing `mouse eat' or `mice eats'\footnote{These are grammatically incorrect, but here we can imagine that the user has already supplied these initial words and just wishes to know the last word.}, with `cheese', which is an output that we can imagine a user being satisfied with.
The model has generalized to form new, desirable predictions.

But consider instead the following:  
\begin{equation}
\label{eq:siblings}
\begin{tabular}{r l l}
  Jack & Jill & siblings \\
  Jane & John & siblings
\end{tabular}  
\end{equation} 
We can imagine these being RDF triples (in subject-object-relation format) or sentences in another language.
Mathematically, (\ref{eq:siblings}) has the same structure as (\ref{eq:cheese}), but the artifacts would now be sentences of the form `Jack John siblings' and `Jane Jill siblings' which are not implied by (\ref{eq:siblings}).
If Jack and John are father and son, say, or unrelated, these would be considered hallucinations. 

Clearly, whether artifacts are generalizations or hallucinations is subjective.
There is no way for the model to differentiate between artifacts that happen to agree with some external standard, and artifacts that do not.
Mathematically, their structures are the same. 
More provocatively, we might say, `generalization is hallucination'.

The only way these artifacts can be precluded is if the dataset contains additional data that contradicts them, such as sentences of the form `Jack John friends' or `Jane Jill unrelated'.
In a large corpus, we might expect that there might be such sentences for \emph{many} common situations, but we certainly cannot hope that we have them for \emph{all}.
We think it is fair to say that artifacts, whether generalizations or hallucinations, are unavoidable.

\subsection{Analogy with recommendation systems}
The field of recommendation systems provides us with a useful analogy.
In this setting, we have a matrix of user-item ratings with many missing entries, and a 
popular method for making recommendations is to find low-rank matrix completions for this matrix \cite{koren2009matrix}.
The point of the low-rank completion is precisely to produce new artifacts, or out of sample predictions.
But anyone who has ever browsed online platforms and wondered, `Why are they recommending this to me?' 
has experienced first-hand how these recommendations are hallucinations in all but name.
One might say that `recommendation is hallucination'.


\subsection{Other types of hallucinations}
We caveat that completion artifacts do not explain all hallucinations, only those commonly called factual fabrications \cite{huang2024survey} or confabulations \cite{sui2024confabulation}: new sentences that the model predicts with high probability but that are unsupportable or false (by an external standard).

Other sources of hallucination include under-fitting (failing to fit to the training fibers) or over-fitting (fitting too well to the pre-training, fine-tuning or RLHF data and thus replicating biases or inaccuracies contained therein) \cite{huang2024survey}.
These hallucinations concern only the training fibers (Fig.~\ref{fig:tensors}b).
Our framework explains hallucinations that occur as artifacts in the `blank space' surrounding those fibers, and suggests that overfitting (usually associated with more parameters) could result in \emph{fewer} artifacts, as the model learns a tensor that is closer to Fig.~\ref{fig:tensors}a rather than Fig.~\ref{fig:tensors}c.

\subsection{Generalization error and overfitting}
The reduction in artifacts when overfitting also explains why generalization error increases when overfitting.
One hopes that a model trained on $\D_{train}$ would produce enough artifacts to include a disjoint dataset $\D_{test}$, and this is less likely when there are fewer artifacts.

Informally, we also expect generalization error to be low if $\D_{test}$ is `in the span' of $\D_{train}$ in some sense.
Our framework lets us formalize this:
let the \emph{effective rank} of a dataset $\D$ be the smallest rank of a tensor that is consistent with $\D$.
Then, low generalization error is more likely if the effective ranks of $\D_{train} \cup \D_{test}$ and $\D_{train}$ are close.

\subsection{Mitigating hallucinations}
Our framework suggests an approach to mitigating hallucinations that arise as artifacts:
in addition to the usual fibers in Fig.~\ref{fig:tensors}b, we need to include the surrounding space in the training data as well.

One method would be to modify the loss function (\ref{eq:loss}) to include a penalty for high-probability predictions outside of the training data.
Another would be to include a special token $t_u$ (for `unsupported') and augment the dataset with sentences\footnote{In database theory, these are \emph{local closed world assumptions} \cite{denecker2008towards}, assertions that statements not in a database are false.
Note that although \cite{cohen2024don} also introduces a special ``I don't know'' token, they ultimately still only use $\D^{n-1}$ in their training, not the rest of $\T^{n-1}$.} $(\t, t_u)$ for a few $\t \in \T^{n-1} \backslash \D^{n-1}$.
Either way, we would need to decide which subset of $\T^{n-1} \backslash \D^{n-1}$ to include in the loss function or as additional data, as using all of $\T^{n-1} \backslash \D^{n-1}$ is impractical.
We could take a random sample, or meticulously curate a subset that we wish to exclude.

\subsection{The generalization-hallucination trade-off}
However, if we take seriously the assertion that `generalization is hallucination', we need to be careful when adopting strategies that mitigate hallucinations arising from artifacts, as they might impact desirable generalizations as well.
Conversely, efforts to improve the generalization capabilities of language models should also study the impact on hallucinations.
Neither should be studied in insolation without considering the trade-off between generalization and hallucination.

\subsection{Relation to other work}
\label{sec:related_work}
Phrases equating generalization and hallucination or claiming that they are two sides of the same coin have been floating around the machine learning community, and are in some sense vacuously true if we define `generalization' and `hallucination' to both just mean `predictions outside the training set'.
However, to our knowledge, there has been no prior work investigating the consequences of this claim, or identifying mechanisms that implicate both phenomena.

The role of compression in generalization has been studied in \cite{arora2018stronger,aghajanyan2020intrinsic,lotfi2022pac,lotfi2023non}, supporting the notion that `simpler descriptions generalize better' \cite{lotfi2022pac}.
Hallucinations are not discussed.
It would be interesting to relate the measures used in those papers, such as intrinsic dimensionality \cite{aghajanyan2020intrinsic}, to tensor rank to see if they account for the same types of generalization behaviors.

The inevitability of hallucinations has been proven in \cite{banerjee2024llms,xu2024hallucination} using computability theory, although their arguments apply more generally to much larger classes of machines and do not provide specific insights on language models. 
Our framework draws the narrative a little more closely around language models, while also including generalizations.

To our knowledge, only \cite{li2024banishing} considers hallucinations and generalization together, finding that  models can have low generalization error but still hallucinate.
Our framework suggests that this should be expected, and we hope that our work can similarly provide a theoretical context for interpreting empirical findings in other papers.

\section{Limitations and future directions}
\label{sec:limitations_and_future_directions}
This is a short opinion paper, and we have not provided theoretical proofs or substantial empirical results to support our claims.
In this section, we list the limitations of this paper, which double as suggestions for future research.

\begin{enumerate}
  \item Although we provide experiments on toy models and datasets, it is uncertain if the same trends will persist when scaled up to actual large language models and large textual datasets. Future research could look into quantifying the number of artifacts present in larger models.
  \item The theoretical objects we define in this paper (such as tensors associated to datasets and to language models, completion artifacts and (effective) rank of tensors) very quickly grow impractical to compute or quantify.
  When the context and vocabulary of a language model gets large, it is impossible to count the number of artifacts as we have done in Figs.~\ref{fig:new_triples_vs_num_triples} and \ref{fig:new_triples_vs_num_params}, as that would entail evaluating the model on every possible $n$-gram.
  Work will have to be done to find ways to approximate or estimate these objects and quantities tractably.
  \item Once the previous two points have been resolved, and we have a way of quantifying artifacts and the ranks of datasets and language models, we can begin to test the claims made in Section \ref{sec:generalization_or_hallucination}, such as:
  \begin{enumerate}[a)]
    \item Overfitting leads to fewer artifacts;
    \item Generalization error increases when overfitting, due to the lack of artifacts;
    \item Generalization error will be low when the effective rank training and testing data is close to the effective rank of training data alone;
    \item Augmenting training data with `unsupported tokens' on sentences outside the training data can help to mitigate hallucinations.
  \end{enumerate}
  \item A more thorough review of the literature can be undertaken to interpret other empirically-observed phenomena in the framework of tensor completions.
\end{enumerate}




\bibliographystyle{plain}
\bibliography{biblio}

\end{document}